\documentclass[journal]{IEEEtran}

\usepackage{amsmath, amssymb, amsfonts}
\usepackage{graphicx}
\usepackage{booktabs}
\usepackage{array}
\usepackage{url}
\usepackage{cite}
\usepackage{textcomp}
\usepackage[T1]{fontenc}
\usepackage[utf8]{inputenc}
\usepackage{threeparttable}
\usepackage{amsopn}
\usepackage[hidelinks, breaklinks]{hyperref}

\begin{document}

\title{Dual-Stream Cross-Anchor Correction: Grounding Long-Form Captions\\and the Domain Limits of Object-Level Anchors}

\author{Lingkai~Bu,
Qian~Gao,
Jun~Fan,
Guohui~Ding,
Zhenyu~Yang,
Yuteng~Xiao,
and~Jinyi~Liang%
\thanks{This work was funded by the projects ZR2022MF333 and ZR2024QF053
supported by Shandong Provincial Natural Science Foundation, in part by the
Pilot Project for Integrated Innovation of Science, Education, and Industry of
Qilu University of Technology (Shandong Academy of Sciences) 2026ZDCX01, and in
part by the projects of the Key Laboratory of Computing Power Network and
Information Security, Ministry of Education under Grant No. 2023ZD028 and Grant
No. 2024PY025. \emph{(Corresponding author: Qian Gao, e-mail:
gq@qlu.edu.cn.)}}%
\thanks{L. Bu, Q. Gao, Z. Yang, and Y. Xiao are with the Key Laboratory of
Computing Power Network and Information Security, Ministry of Education,
Shandong Computer Science Center (National Supercomputer Center in Jinan),
Qilu University of Technology (Shandong Academy of Sciences), Jinan 250014,
Shandong, China; the Shandong Engineering Research Center of Big Data Applied
Technology, Faculty of Computer Science and Technology, Qilu University of
Technology (Shandong Academy of Sciences), Jinan 250353, Shandong, China; and
the Shandong Provincial Key Laboratory of Industrial Network and Information
System Security, Shandong Fundamental Research Center for Computer Science,
Jinan 250014, Shandong, China.}%
\thanks{J. Fan is with China Telecom Digital Intelligence Technology Co., Ltd.,
No. 1999 Shunhua Road, Jinan 250101, Shandong, China.}%
\thanks{G. Ding is with Shenyang Aerospace University, Shenyang, China.}%
\thanks{J. Liang is with the University of Jinan, Jinan 250002, Shandong,
China.}%
\thanks{The source code is available at
\url{https://github.com/lingkaibu/DSCC}.}}

\maketitle

\begin{abstract}
Object hallucination in multimodal large language models arises when language priors and corpus co-occurrence bias outweigh the visual evidence, with nothing tying an object mention to the image. Most remedies intervene at decoding time, yet under a unified protocol their benefit is confined to short captions; supervised fine-tuning (SFT) on a detail-rich corpus lengthens captions, but over forty percent still name absent objects. This paper proposes Dual-Stream Cross-Anchor Correction (DSCC). Unlike work that post-processes decoding, DSCC injects object-level visual anchors into the language model itself during fine-tuning: a perception stream aligns object-level hidden states at an intermediate layer to frozen text anchors by a bidirectional contrastive objective; a cognition stream lets deeper layers query those anchors by cross-attention at every generation step; and a two-stage curriculum gate couples them, making evidence retrieval a structural constraint on generation. Under one backbone and one scoring protocol, experiments span long-caption hallucination, object-existence discrimination and cross-domain generalisation, with vanilla SFT on the same corpus and schedule as a length- and density-matched control separating the data effect from the architectural gain. DSCC alone reaches the long-caption, low-hallucination region: captions roughly 1.9 times the baseline length at 88.19\% precision per object mention, the highest under a density-independent criterion. Ablations expose a synergy: the perception stream alone degrades precision yet reverses sign when stacked on the cognition stream. No universal superiority is claimed: three out-of-domain benchmarks yield a predictable, falsifiable domain-conditionality, the synergy being bound to the anchors' semantic domain and breaking on charts and illusions.
\end{abstract}


\IEEEpeerreviewmaketitle

\section{Introduction}
\label{sec:1}

Multimodal large language models (MLLMs) have shown remarkable understanding and generation ability across general vision-language tasks \cite{r1,r2,r3}, yet suffer from object hallucination, that is, they confidently describe objects that the image does not contain \cite{r4,r5}, a flaw that directly obstructs the deployment of MLLMs wherever reliability matters. Given a kitchen photograph containing only a sink and an oven, LLaVA-1.5 writes a refrigerator, bottles, cups, knives and a bowl into its description, five objects that are simply not present.

Three properties make hallucination a problem to be solved rather than a stylistic blemish to be tolerated. The first is concealment: what gets invented is usually the object that co-occurs most strongly with the scene and therefore reads as reasonable, a refrigerator in a kitchen or cutlery on a dining table, so from the text alone a reader can hardly tell which clause is grounded. The second is propagation: in a deployed system downstream modules consume the generated text rather than the original image, so once a fabricated object enters the caption it travels onward as established fact, and no later stage typically returns to the pixels to check. The third is that severity is set by the application rather than by the number of errors: in a high-stakes setting a single fabricated object suffices to do real damage. In accessible image description for blind and low-vision users no second channel exists against which the caption can be verified, so a spoken phantom object turns into misleading guidance and, in the worst case, physical risk \cite{r5}. Captions produced by MLLMs are also widely reused as visual instruction data, ShareGPT4V \cite{r6} being a prominent case, so unverified hallucinations are recycled into the training corpus of the next generation and the error is amplified across iterations. These concerns converge on one requirement: every object a model asserts should rest on verifiable visual evidence, the more pressing the longer and more detailed the caption, since fabrications are then harder to spot.

Efforts to mitigate hallucination have largely been trapped by an implicit length-quality trade-off: they never leave the short-caption regime, and within that regime the improvement quickly saturates. Under a unified protocol (the same LLaVA-1.5-7B backbone, the same 500 COCO \cite{r49} images, the same scorer), the intervention-free baselines and training-free decoding-time methods such as VCD \cite{r7} and OPERA \cite{r8} all produce captions within a narrow band of roughly 90--105 words, and the strongest of them lowers the sentence-level hallucination rate without lengthening the caption but goes no further. Standard supervised fine-tuning on a detail-rich corpus such as ShareGPT4V \cite{r6} has the opposite character: it elicits long captions, yet lacking any visual grounding constraint it leaves the absolute hallucination rate high; under the same protocol (setup in Section~\ref{sec:4.1}), Section~\ref{sec:4.2.2} shows that 41.60\% of the captions still contain a hallucinated object. Research has therefore left one question largely unanswered: how can a model say more while getting less wrong?

This paper answers that question with Dual-Stream Cross-Anchor Correction (DSCC). Rather than intervening at decoding time, DSCC moves the grounding constraint into fine-tuning, inside the language model: fine-grained anchors are built at an intermediate layer for the objects present in the image, and the deeper layers query them at every forward step of generation, so that searching for visual evidence becomes a structural constraint rather than a probability correction after the fact.

The main contributions of this paper are as follows.

(1) A new perspective. Comparisons among hallucination mitigation methods have long been confined to the short-caption regime of roughly 90--105 words. The link between caption length and hallucination is not itself a new topic: statistical analyses have observed that hallucinations concentrate in the later part of the generated text, which motivated a post-hoc revision model \cite{r9}, and a recent study of why longer responses hallucinate more attributes the risk to a growing reliance on context rather than to length as such \cite{r10}. What differs here is that caption length and object density are promoted from confounders explained away after the fact to coordinates that must be reported alongside the hallucination rate, and a control matched in both length and object density is set up accordingly, instead of comparing a single hallucination number.

(2) A mechanism finding. At the component level the dual-stream design is a controlled combination of existing building blocks: the cognition stream is gated cross-attention in the manner of Flamingo, with its key/value source replaced by perception anchors internal to the language model; the perception stream is region-level alignment in the manner of RegionCLIP pushed into the language model; the curriculum gate is a standard warmup schedule. The contribution lies not in the blocks but in the mechanism their combination reveals: the division of labour between the two streams, the interaction whereby the perception stream is harmful alone yet turns synergistic once combined, and the predictable domain-conditionality that this synergy inherits from the CLIP-COCO anchor binding.

(3) Empirical insights. Standard supervised fine-tuning on the same corpus with the same schedule serves as a controlled reference, separating the effect of the data paradigm, namely that captions become longer, from the net gain of the dual-stream architecture itself. This paper further proposes the domain-conditionality of a hallucination mitigation mechanism: the range over which a mechanism is effective can be delimited in advance by the semantic domain of the anchors it relies on, making that range verifiable and falsifiable.

\section{Related Work}
\label{sec:2}

Several largely independent lines of work address object-level hallucination in MLLMs. The first is training-free intervention at the decoding stage: VCD \cite{r7}, OPERA \cite{r8}, DoLa \cite{r11}, HALC \cite{r12}, M3ID \cite{r13}, ICD \cite{r14}, SID \cite{r15}, CCA \cite{r16} and AGLA \cite{r17}. These methods are plug-and-play and need no retraining, but act on surface symptoms in the output probabilities or the attention distribution, and have so far been verified mainly in the short-caption regime (Section~\ref{sec:4.3}). The second line is post-hoc refinement: Woodpecker \cite{r18}, LURE \cite{r9}, Volcano \cite{r19}, HalluciDoctor \cite{r20} and LogicCheckGPT \cite{r21} repair the initial output from outside the forward pass, usually with an extra detector, an external LLM, or multi-round inference; they are treated as a direction orthogonal to DSCC, since a post-hoc strategy can be stacked on the long outputs of DSCC and is left as future work (Section~\ref{sec:6}). Post-hoc refinement is therefore excluded from the main comparison, a deliberate and openly stated scoping decision rather than an attempt to avoid a strong baseline. The third line is preference optimisation and alignment: building on RLHF \cite{r22} and DPO \cite{r23}, RLHF-V \cite{r24}, LLaVA-RLHF \cite{r25}, mDPO \cite{r26} and CSR \cite{r27} numerically suppress the generation probability of hallucinatory words. The common limitation is that this remains an implicit reshaping of probabilities in text space: no structural hard constraint is established between image features and linguistic symbols. The fourth line is contrastive grounding: contrastive representation learning runs from MoCo \cite{r28} through CLIP \cite{r30}, ALIGN \cite{r31}, BLIP \cite{r32} and SigLIP \cite{r33}, while GLIP \cite{r34}, RegionCLIP \cite{r35} and Grounding DINO \cite{r36} push the alignment signal towards region-phrase granularity and BLIP-2 \cite{r37} bridges a frozen visual encoder and an LLM through a Q-Former; in current MLLM architectures, however, such alignment mostly stops at the visual encoder's output and does not reach inside the language model. On backbones, Flamingo \cite{r38}, LLaVA \cite{r39}, LLaVA-1.5 \cite{r1}, MiniGPT-4 \cite{r2}, InstructBLIP \cite{r3} and mPLUG-Owl2 \cite{r40} have established the dominant paradigm of a visual encoder, typically ViT-style \cite{r41}, a projector and an LLM; LLaVA-1.5-7B \cite{r1} is the backbone here. On evaluation, object-level hallucination is measured chiefly by POPE \cite{r4} and CHAIR \cite{r5}; MME \cite{r42} lies within the COCO object semantic domain and serves as an out-of-distribution but same-domain test, whereas HallusionBench \cite{r48} and MMHal-Bench \cite{r25}, covering charts and optical illusions and abstract scenes, serve as a genuinely out-of-domain boundary test (Section~\ref{sec:4.5}). In sum, existing methods either operate on output probabilities and text or keep the grounding constraint on the vision-encoder side, whereas DSCC injects it during training, inside the language model, at every generation step, while remaining compatible with, and stackable on, the decoding-time methods above.

\section{Method}
\label{sec:3}

\subsection{Overview}
\label{sec:3.1}

Dual-Stream Cross-Anchor Correction (DSCC), proposed in this study, is a fine-tuning framework. On top of standard supervised fine-tuning (SFT) over detail-rich caption data, it introduces two auxiliary streams that suppress hallucination at the perceptual level, namely which objects are claimed to exist in the image, and at the cognitive level, namely how the content generated in the deep layers drifts away from the visual evidence. DSCC is designed for the following situation: the MLLM has already been fine-tuned to produce long, object-dense captions, so that the central challenge is not to shorten the output but to raise the precision of each mention without sacrificing coverage.

Guided by the empirical finding that the shallow Transformer layers of an MLLM mainly encode visual perception while the deep layers perform language-level reasoning, DSCC grafts three coupled components onto a pretrained MLLM backbone (LLaVA-1.5-7B is used here \cite{r1}), all of which act additively on top of the standard SFT loss:

(1) Perception stream (Section~\ref{sec:3.2}): a fine-grained, object-level contrastive objective that anchors shallow visual representations to CLIP-aligned text semantics, attacking perceptual hallucination head on.

(2) Cognition stream (Section~\ref{sec:3.3}): a cross-attention module that queries the perception anchors as evidence at every generation step of the deep layers, structurally severing the path along which deep reasoning drifts from the visual evidence.

(3) Curriculum-gated fine-tuning, CGFT (Section~\ref{sec:3.4}): a two-stage schedule that couples the two streams progressively, letting the perception stream establish stable anchors before the cognition stream queries them.

\begin{figure*}[!t]
\centering
\includegraphics[width=0.88\textwidth]{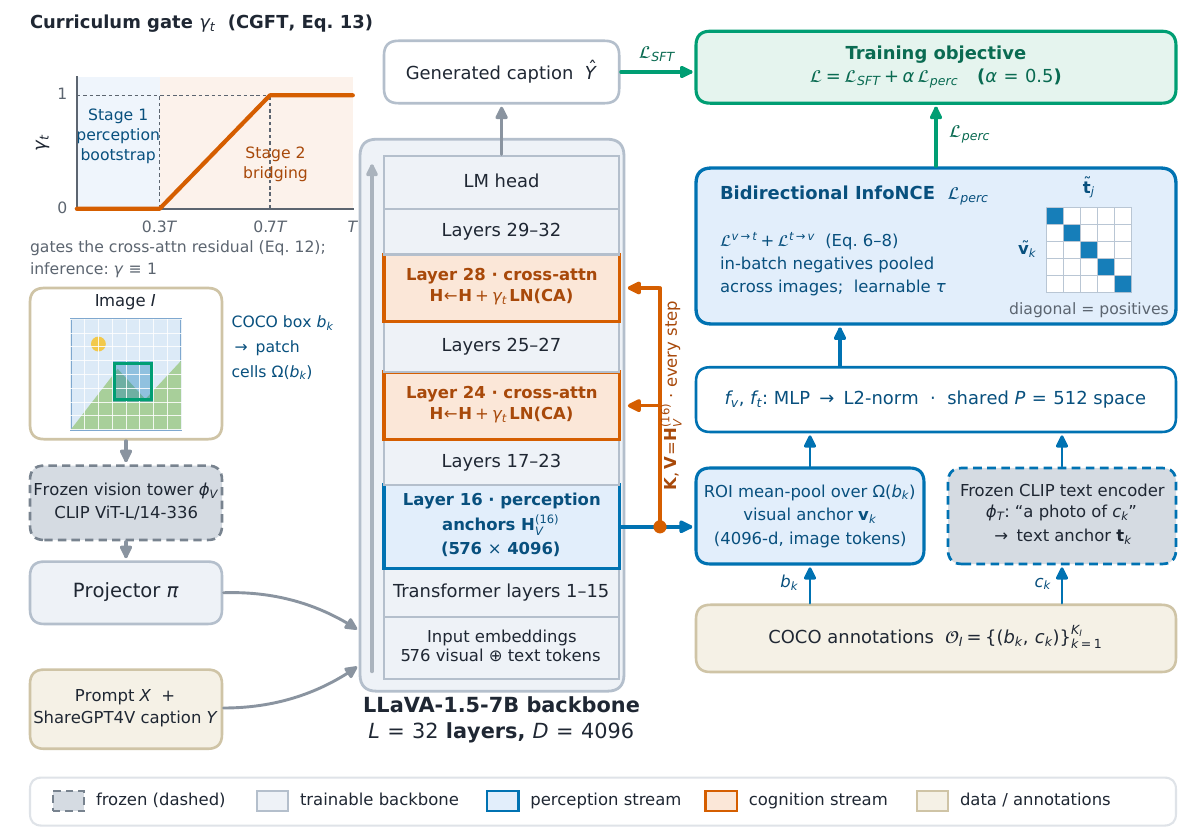}
\caption{Overview of the DSCC architecture. The perception stream aligns ROI features at layer 16 to frozen CLIP text anchors through an object-level InfoNCE loss; the cognition stream injects gated cross-attention at layers 24 and 28, taking the hidden states of all image tokens at layer 16 as keys and values at every generation step; a two-stage curriculum gate $\gamma$\_t couples the two streams progressively.}
\label{fig:2}
\end{figure*}

Two design choices set DSCC apart from existing hallucination mitigation methods. First, what is modified is the model itself rather than the decoding rule, so the grounding path is active at every generation step instead of at decoding time alone. Second, the dual-stream modules act strictly as auxiliaries to the language modelling backbone: only \(\mathcal{L}_{\mathrm{SFT}}\) determines the output distribution (length, object density, coverage), while the perception and cognition streams contribute refinement signals solely through \(\mathcal{L}_{\mathrm{perc}}\) \eqref{eq:8} and gated residual injection \eqref{eq:12}. The vision tower stays frozen throughout; all remaining parameters are optimised jointly. Fig.~\ref{fig:2} gives an overview of the resulting architecture.

Every symbol used in \eqref{eq:2}--\eqref{eq:14} is defined in Table~\ref{tab:0}, together with the default value of each hyperparameter; the training setup in which those values are used is given in Section~\ref{sec:4.1.1}.

\begin{table}[!t]
\centering
\caption{Notation of (1)--(14) and default hyperparameters.}
\label{tab:0}
\footnotesize
\setlength{\tabcolsep}{3pt}
\begin{tabular}{@{}lp{0.46\columnwidth}l@{}}
\toprule
Symbol & Meaning & Value\\
\midrule
\(N\), \(G\), \(D\) & Image tokens, grid side, width & 576, 24, 4096\\
\(\mathbf{H}^{(l)}\), \(\mathbf{H}_{V}^{(l)}\) & Layer-\(l\) states; image only & ---\\
\(b_{k}\), \(c_{k}\), \(\Omega(b_{k})\) & Box, class name, patch cells & COCO\\
\(\mathbf{v}_{k}\), \(\mathbf{t}_{k}\), \(\phi_{T}\) & Visual, text anchor; encoder & ---\\
\(f_{v}\), \(f_{t}\), \(P\) & Projection MLPs; dimension & 512\\
\(\tau\), \(K\) & Temperature; objects/batch & 0.07 init.\\
\(\alpha\) & Weight of \(\mathcal{L}_{\mathrm{perc}}\) & 0.5\\
\(\mathbf{W}^{Q,K,V}\), \(\mathbf{W}^{O}\) & Cross-attention projections & ---\\
\(h\), \(d_{h}\) & Heads; head width & 32, 128\\
\(\gamma_{t}\), \(T\) & Curriculum gate; total steps & \(0.3T\)--\(0.7T\)\\
\(l_{p}\), \(\mathcal{L}_{c}\) & Perception; cognition layers & 16, \(\{ 24,28\}\)\\
\bottomrule
\end{tabular}
\end{table}

\subsection{Perception Stream: Fine-Grained Object Contrast}
\label{sec:3.2}

\subsubsection{Object-Level Visual Anchors}
\label{sec:3.2.1}

Sentence-level vision-language contrast such as CLIP is too coarse to ground an individual object, so per-object visual anchors are extracted instead, by projecting each ground-truth bounding box onto the patch grid.

Consider an image of size \((W,H)\) that contains an object with bounding box \(b_{k} = (x_{k},y_{k},w_{k},h_{k})\), indexed by \(k\). Its discrete grid coverage on the \(G \times G\) patch grid is defined as:

\begin{equation}
\label{eq:2}
\Omega(b_{k}) = \left\{ (i,j) \in [0,G)^{2} \;\middle|\; \begin{aligned}&i_{\min} \leq i < i_{\max},\\ &j_{\min} \leq j < j_{\max}\end{aligned} \right\}
\end{equation}

where

\begin{equation}
\label{eq:2b}
\begin{aligned}
i_{\min} &= \min\!\left(G-1,\ \left\lfloor \tfrac{x_{k}}{W}G \right\rfloor\right)\!,\\
i_{\max} &= \max\!\left(i_{\min}+1,\ \min\!\left(G,\ \left\lceil \tfrac{x_{k}+w_{k}}{W}G \right\rceil\right)\right)\!,\\
j_{\min} &= \min\!\left(G-1,\ \left\lfloor \tfrac{y_{k}}{H}G \right\rfloor\right)\!,\\
j_{\max} &= \max\!\left(j_{\min}+1,\ \min\!\left(G,\ \left\lceil \tfrac{y_{k}+h_{k}}{H}G \right\rceil\right)\right)\!.
\end{aligned}
\end{equation}

Here \((i,j)\) indexes the columns and rows of the patch grid and \(\lfloor \cdot \rfloor\), \(\lceil \cdot \rceil\) are the floor and ceiling functions, so \eqref{eq:2b} converts the pixel coordinates of \(b_{k}\) into patch units. Clamping to \(\lbrack 0,G)\) keeps rounded box coordinates inside the grid, while the bound \(i_{\max} \geq i_{\min} + 1\) (\(j\) likewise) enforces \(|\Omega(b_{k})| \geq 1\), so that a box smaller than a single patch still contributes exactly one anchor patch instead of being silently dropped.

The visual anchor of object \(k\) is the mean of the hidden states at the perception layer over the patches in \(\Omega(b_{k})\):

\begin{equation}
\label{eq:3}
\mathbf{v}_{k} = \frac{1}{|\Omega(b_{k})|}\sum_{(i, j) \in \Omega(b_{k})}^{}\mathbf{H}_{V}^{(l_{p})}\left\lbrack i \cdot G + j \right\rbrack \in \mathbb{R}^{D}
\end{equation}

where \(\mathbf{H}_{V}^{(l_{p})}\lbrack i \cdot G + j \rbrack\) is the hidden state of the image token at grid cell \((i,j)\), so that the sum divided by \(|\Omega(b_{k})|\) mean-pools the tokens inside the box alone. Because \(\mathbf{H}_{V}^{(l_{p})}\) is restricted to image-token positions, \(\mathbf{v}_{k}\) is purely visual, free of contamination by prompt or reply tokens, which removes the self-loop risk that arises when the entire sequence is pooled.

\subsubsection{Object-Level Text Anchors}
\label{sec:3.2.2}

To match the vision-language semantic space, the frozen CLIP text encoder paired with \(\phi_{V}\) during CLIP pretraining, that is \(\phi_{T}\), is used instead of a context-independent lookup in the LLM input embedding table:

\begin{equation}
\label{eq:4}
\mathbf{t}_{k} = \phi_{T}\!\left(\text{``a photo of ''} + c_{k}\right) \in \mathbb{R}^{d_{t}}
\end{equation}

The ``a photo of'' template follows the standard CLIP zero-shot recipe.

\subsubsection{Bidirectional InfoNCE Objective}
\label{sec:3.2.3}

Both kinds of anchor are projected by lightweight MLPs into a shared space of dimension \(P\) (\(P = 512\)) and L2-normalised:

\begin{equation}
\label{eq:5}
{\widetilde{\mathbf{v}}}_{k} = \frac{f_{v}(\mathbf{v}_{k})}{\| f_{v}(\mathbf{v}_{k}) \|_{2}},{\widetilde{\mathbf{t}}}_{k} = \frac{f_{t}(\mathbf{t}_{k})}{\| f_{t}(\mathbf{t}_{k}) \|_{2}}
\end{equation}

where the division by the Euclidean norm places every anchor on the unit sphere of \(\mathbb{R}^{P}\), so that the inner products below are cosine similarities. Over all \(K = \sum_{b}^{}K_{I_{b}}\) objects pooled from a training batch, where objects from different images serve as in-batch negatives for one another, a symmetric InfoNCE loss is applied:

\begin{equation}
\label{eq:6}
\mathcal{L}_{\mathrm{perc}}^{v \rightarrow t} = - \frac{1}{K}\sum_{k = 1}^{K}\log\frac{\exp({\widetilde{\mathbf{v}}}_{k}^{\top}{\widetilde{\mathbf{t}}}_{k}/\tau)}{\sum_{j = 1}^{K}\exp({\widetilde{\mathbf{v}}}_{k}^{\top}{\widetilde{\mathbf{t}}}_{j}/\tau)}
\end{equation}

\begin{equation}
\label{eq:7}
\mathcal{L}_{\mathrm{perc}}^{t \rightarrow v} = - \frac{1}{K}\sum_{k = 1}^{K}\log\frac{\exp({\widetilde{\mathbf{v}}}_{k}^{\top}{\widetilde{\mathbf{t}}}_{k}/\tau)}{\sum_{j = 1}^{K}\exp({\widetilde{\mathbf{v}}}_{j}^{\top}{\widetilde{\mathbf{t}}}_{k}/\tau)}
\end{equation}

\begin{equation}
\label{eq:8}
\mathcal{L}_{\mathrm{perc}} = \frac{1}{2}\left( \mathcal{L}_{\mathrm{perc}}^{v \rightarrow t} + \mathcal{L}_{\mathrm{perc}}^{t \rightarrow v} \right)
\end{equation}

In \eqref{eq:6} and \eqref{eq:7} the numerator scores the matched vision-text pair of the same object, \(({\widetilde{\mathbf{v}}}_{k},{\widetilde{\mathbf{t}}}_{k})\), while the denominator sums over all \(K\) candidates: \eqref{eq:6} contrasts one visual anchor against every text anchor in the batch and \eqref{eq:7} does the reverse, and \eqref{eq:8} averages the two directions. The temperature \(\tau\) is learnable, initialised at 0.07 following CLIP practice, and clamped at \(\log\tau^{- 1} \leq \log100\) to prevent numerical divergence.

\subsection{Cognition Stream: Cross-Anchor Attention Injection}
\label{sec:3.3}

\subsubsection{Anchor-Conditioned Generation}
\label{sec:3.3.1}

The cognition stream is required to attend to the perceptual evidence at every generation step. At each cognition layer \(l \in \mathcal{L}_{c}\), a multi-head cross-attention module~\cite{r50} is inserted that takes the cognitive hidden state as query and the hidden states of all image tokens at the perception layer, hereafter the perception anchors, as key and value:

\begin{equation}
\label{eq:9}
\begin{split}
&\operatorname{CrossAttn}^{(l)}\!\left( \mathbf{H}^{(l)},\mathbf{H}_{V}^{(l_{p})} \right)\\
&\qquad = \operatorname{Concat}\!\left\lbrack \mathbf{O}_{1}^{(l)};\ldots;\mathbf{O}_{h}^{(l)} \right\rbrack\mathbf{W}_{l}^{O}
\end{split}
\end{equation}

Each head is computed by scaled dot-product attention:

\begin{equation}
\label{eq:10}
\mathbf{O}_{i}^{(l)} = \operatorname{softmax}\left( \frac{\mathbf{Q}_{i}^{(l)}(\mathbf{K}_{i}^{(l)})^{\top}}{\sqrt{d_{h}}} \right)\mathbf{V}_{i}^{(l)}
\end{equation}

\begin{equation}
\label{eq:11}
\begin{split}
\mathbf{Q}_{i}^{(l)} &= \mathbf{H}^{(l)}\mathbf{W}_{l,i}^{Q},\qquad
\mathbf{K}_{i}^{(l)} = \mathbf{H}_{V}^{(l_{p})}\mathbf{W}_{l,i}^{K},\\
\mathbf{V}_{i}^{(l)} &= \mathbf{H}_{V}^{(l_{p})}\mathbf{W}_{l,i}^{V}.
\end{split}
\end{equation}

Here the cognitive hidden state supplies the queries while the perception anchors supply the keys and values \eqref{eq:11}, \(\sqrt{d_{h}}\) scales the dot product before the softmax \eqref{eq:10}, and \(\mathbf{W}_{l}^{O}\) recombines the \(h\) head outputs \eqref{eq:9}; these projections are the only parameters added to the backbone. The module uses \(h = 32\) heads with head dimension \(d_{h} = D/h = 128\), exactly the configuration used by LLaMA internally, which keeps bf16 attention numerically stable and lets the cross-attention maps be read in the same per-head framework as the self-attention of the language model.

\subsubsection{Residual Injection with Near-Identity Initialisation}
\label{sec:3.3.2}

The cognitive hidden state is updated by a gated residual:

\begin{equation}
\label{eq:12}
\begin{split}
\mathbf{H}^{(l)} \leftarrow{}& \mathbf{H}^{(l)} + \gamma_{t} \cdot \operatorname{LN}\!\left( \operatorname{CrossAttn}^{(l)}\!\left(\mathbf{H}^{(l)},\mathbf{H}_{V}^{(l_{p})}\right) \right)\!,\\
& l \in \mathcal{L}_{c}.
\end{split}
\end{equation}

where \(\gamma_{t} \in \lbrack 0,1\rbrack\) is the curriculum gate (Section~\ref{sec:3.4}) and LN denotes LayerNorm, the residual form ensuring that the injected branch adds to the stream instead of replacing it. The output projection \(\mathbf{W}_{l}^{O}\) is initialised from \(\mathcal{N}(\mathbf{0},\sigma^{2}\mathbf{I})\) with \(\sigma = 10^{- 3}\), so the initial contribution of the branch, about \(\gamma_{t} \cdot \mathcal{O}(\sigma\sqrt{D}) \approx \gamma_{t} \cdot 0.082\), is small relative to the \(\mathcal{O}(1)\) residual stream: the near-identity behaviour named in the heading. Exactly \(\mathbf{W}_{l}^{O} = \mathbf{0}\) is avoided, as it would leave the attention projections without gradient (supplementary material).

Why cross-attention rather than post-hoc alignment? A naive alternative is to minimise \(\| \mathbf{H}^{(L)} - \mathbf{H}^{(l_{p})} \|^{2}\) directly. That would violate the principle of hierarchical abstraction: the deep layers are supposed to encode semantics different from the shallow ones, and forcing them to coincide collapses the hierarchy of representations. Cross-attention instead lets the cognition stream query the perception stream during generation, which introduces the grounding constraint while preserving the natural depth-wise specialisation.

\subsection{Curriculum-Gated Fine-Tuning (CGFT)}
\label{sec:3.4}

A two-stage curriculum couples the two streams progressively. Let \(T\) denote the total number of training steps and \(t \in \lbrack 0,T\rbrack\) the current step. In both stages the model is trained under the single objective \(\mathcal{L}_{\mathrm{SFT}} + \alpha\mathcal{L}_{\mathrm{perc}}\), formalised as \eqref{eq:14} in Section~\ref{sec:3.5}; the curriculum modulates only the cognition-perception coupling gate \(\gamma_{t}\) inside the cross-attention residual \eqref{eq:12}.

Stage 1: perception bootstrap (\(0 \leq t < 0.3T\))

The cognition gate is closed, \(\gamma_{t} = 0\). Cross-anchor attention contributes nothing to the residual stream, so the deep layers behave exactly as in vanilla LLaVA and gradients flow only through \(\mathcal{L}_{\mathrm{SFT}}\) and \(\mathcal{L}_{\mathrm{perc}}\). This stage establishes object-level vision-text anchors at \(l_{p}\) before the deep layers are perturbed, echoing the pretrain-then-finetune inductive bias.

Stage 2: cognition-perception bridging (\(0.3T \leq t \leq T\))

The gate rises linearly and then saturates at 1:

\begin{equation}
\label{eq:13}
\gamma_{t} = \min\left( \frac{t - 0.3T}{0.4T},1 \right) \in \lbrack 0,1\rbrack
\end{equation}

so that with \(t\) the current step and \(T\) the total number of steps the gate opens linearly between \(0.3T\) and \(0.7T\) and stays at 1 thereafter. Once \(\gamma_{t} > 0\), the cognition layers begin to query the perception anchors through cross-attention \eqref{eq:12}, while the perception stream keeps refining those anchors under the same \(\mathcal{L}_{\mathrm{perc}}\); from this point the two streams adapt jointly. The slow ramp prevents an abrupt distribution shift in the cognition layers: under bf16 training, switching instantaneously from \(\gamma = 0\) to \(\gamma = 1\) destabilises the optimisation, as the ablation in Section~\ref{sec:4} shows.

Why a two-stage curriculum rather than single-stage joint training? If cross-attention were enabled from \(t = 0\) onwards, the cognition stream would attend to perception anchors not yet aligned with their CLIP text counterparts; the resulting noisy attention output would backpropagate into the perception stream through \(\mathcal{L}_{\mathrm{SFT}}\), compete with \(\mathcal{L}_{\mathrm{perc}}\) and slow anchor convergence. Stage 1, with the gate closed, lets the perception stream converge in isolation, after which Stage 2 introduces cognition-perception coupling on anchors that have already taken shape.

\subsection{Overall Training Objective}
\label{sec:3.5}

The two streams are combined under the curriculum gate:

\begin{equation}
\label{eq:14}
\mathcal{L}_{\mathrm{total}}(t) = \mathcal{L}_{\mathrm{SFT}} + \alpha\mathcal{L}_{\mathrm{perc}}
\end{equation}

where \(\alpha = 0.5\) balances the supervised long-caption loss against the perception objective of \eqref{eq:8}, and the curriculum schedule enters implicitly, through the gate \(\gamma_{t}\) \eqref{eq:12} inside the forward pass of \(\mathcal{L}_{\mathrm{SFT}}\). All hyperparameters of Table~\ref{tab:0} are held identical across the configurations compared in this paper; the corpus, schedule and optimisation settings under which they are used are reported in Section~\ref{sec:4.1.1}, and the remaining implementation details in the supplementary material.

\section{Experiments}
\label{sec:4}

The experiments are organised around a single claim: the gain of DSCC is not bought by saying less.

\subsection{Experimental Setup}
\label{sec:4.1}

\subsubsection{Model and the Four-Way Comparison}
\label{sec:4.1.1}

All experiments use LLaVA-1.5-7B as backbone and are trained on the same corpus (ShareGPT4V GPT-4V long captions intersected with COCO object annotations, about 95k samples) under the same two-epoch schedule (roughly 25k optimisation steps). The dual-stream hyperparameters are those of Table~\ref{tab:0} and are identical in every configuration, in particular \(l_{p} = 16\), \(\mathcal{L}_{c} = \{ 24,28\}\) and \(\alpha = 0.5\); optimiser, precision and batch settings are listed in Table~S-1 of the supplementary material. To separate the contribution of the dual-stream architecture cleanly from that of the supervision data itself, four checkpoints are trained and compared that differ only in which dual-stream modules are enabled (Table~S-2): D, both streams off, that is vanilla SFT; A, perception stream only; B, cognition stream only; and C, full dual stream.

The four configurations share corpus and step count, so any difference between D and \{A, B, C\} is attributable to the dual-stream modules rather than to a data-induced shift of the output distribution (Section~\ref{sec:3.1}).

The inference-time gate is fixed at $\gamma$ = 1, consistent with the late curriculum, so the grounding path stays active at every generation step.

\subsubsection{Baselines, Benchmarks and Metrics}
\label{sec:4.1.2}

Besides the four configurations, two representative training-free decoding-time methods are compared, VCD \cite{r7} and OPERA \cite{r8}, with beam search and DoLa additionally listed on POPE for reference; post-hoc refinement methods (Woodpecker, LURE) rely on external detectors or LLMs, are orthogonal to DSCC and, following the scoping in Section~\ref{sec:2}, are excluded from the main comparison and left as stackable future work. In-domain evaluation stays inside the COCO object semantic domain and uses two benchmarks. POPE \cite{r4} tests object existence discrimination over three subsets, Random, Popular and Adversarial, each with N = 3000 and a 50/50 balance of positive and negative samples; the Adversarial subset is the hardest and the main object of analysis. CHAIR \cite{r5} measures the hallucination rate of open-ended captions under the standard protocol of OPERA and VCD (val2014 shuffled with seed 42, 500 images). Out of domain the benchmarks are MME-Hallucination \cite{r42}, HallusionBench \cite{r48} and MMHal-Bench \cite{r25} (Section~\ref{sec:4.5}). The reported metrics are accuracy, precision, recall, F1 and YesRatio on POPE, and CHAIR\_S$\downarrow$, CHAIR\_I$\downarrow$, object recall, mean word count \#Words and objects per caption Obj/Cap on CHAIR. The comparison regime is marked under each table: the CHAIR baselines were reproduced here under an identical protocol, whereas the POPE baselines are quoted from the original papers.

\subsubsection{Statement of Protocol Deviation}
\label{sec:4.1.4}

The OOD evaluations use a scoring protocol that differs from the official leaderboards: HallusionBench and MME are scored by plain string yes/no matching rather than the official GPT-4 judge, and MMHal-Bench by gpt-5.4-mini rather than the official GPT-4-0314. Their absolute scores therefore cannot be compared with any leaderboard, no claim of state of the art is made on any OOD benchmark, and every OOD conclusion is restricted to the relative standing of the four configurations under one scorer.

\subsection{Main Results}
\label{sec:4.2}

\subsubsection{POPE Object Existence Discrimination}
\label{sec:4.2.1}

Table~\ref{tab:3} first compares DSCC with representative decoding-time methods on POPE; Table~S-3 and~\ref{tab:5} then give the controlled results for the four configurations.

\begin{table}[!t]
\centering
\caption{POPE comparison with representative decoding-time methods (MSCOCO, LLaVA-1.5-7B backbone, F1$\uparrow$).}
\label{tab:3}
\footnotesize
\setlength{\tabcolsep}{4pt}
\begin{threeparttable}
\begin{tabular}{@{}lccccc@{}}
\toprule
Method & Type & Random & Popular & Adv. & Mean F1\\
\midrule
LLaVA-1.5 (regular)\tnote{$\dagger$} & no interv. & 81.33 & 80.06 & 77.57 & 79.65\\
VCD\tnote{$\dagger$} & decoding & 87.16 & 85.06 & 81.33 & 84.52\\
Beam search\tnote{$\dagger$} & decoding & --- & --- & --- & 84.90\\
DoLa\tnote{$\dagger$} & decoding & --- & --- & --- & 83.20\\
OPERA\tnote{$\dagger$} & decoding & --- & --- & --- & 85.40\\
D, both streams off & training & 88.41 & 87.28 & 83.83 & 86.51\\
\textbf{DSCC (C, ours)} & training & 87.23 & 86.10 & 83.80 & 85.71\\
\bottomrule
\end{tabular}
\begin{tablenotes}[flushleft]\footnotesize
\item[$\dagger$] Quoted from the original papers and not re-run: VCD and its regular baseline from Leng \textit{et al.}; the three-subset mean F1 of beam search, DoLa and OPERA from Huang \textit{et al.}, whose paper does not list the subsets separately (marked ``---''). The backbone is likewise LLaVA-1.5-7B, but off the shelf, that is, without SFT on ShareGPT4V $\times$ COCO.
\end{tablenotes}
\end{threeparttable}
\end{table}

Two points of honest accounting must be stated: First, the gap to the literature baselines also contains the SFT data effect, since D and C were fine-tuned on ShareGPT4V $\times$ COCO whereas those baselines are off the shelf (Section~\ref{sec:4.4.1}). Second, the POPE contribution of DSCC lies in precision rather than F1 (Table~\ref{tab:5}), while most literature baselines report only accuracy and F1. On the comparable F1 criterion, the mean F1 of DSCC (85.7) matches the strongest decoding-time method, OPERA (85.4), and exceeds VCD, beam search and DoLa; on the hardest Adversarial subset it reaches F1 83.8 against VCD's 81.3. For a training-time method that adds no decoding overhead and produces captions about 1.9 times longer, this is a competitive showing, though no claim of state of the art is made on that basis.

Within the controlled four-way comparison (same corpus, same number of steps), the hardest Adversarial subset reveals a clear monotone trend (Table~\ref{tab:5}): precision rises monotonically as the streams are added, A (0.8315) \textless{} D (0.8510) \textless{} B (0.8638) \textless{} C (0.8839), the main model C improving on the vanilla SFT baseline D by 3.3 percentage points.

\begin{table}[!t]
\centering
\caption{Four-way comparison on the POPE Adversarial subset, ordered by precision.}
\label{tab:5}
\footnotesize
\setlength{\tabcolsep}{3pt}
\resizebox{\ifdim\width>\columnwidth \columnwidth\else\width\fi}{!}{%
\begin{tabular}{@{}lccccc@{}}
\toprule
ckpt & Acc & \textbf{Prec} & Recall & F1 & YesRatio\\
\midrule
A, perception only & 0.8330 & 0.8315 & 0.8353 & 0.8334 & 0.5023\\
D, both off & 0.8407 & 0.8510 & 0.8260 & 0.8383 & 0.4853\\
B, cognition only & 0.8420 & 0.8638 & 0.8120 & 0.8371 & 0.4700\\
\textbf{C, full dual stream} & \textbf{0.8460} & \textbf{0.8839} & 0.7967 & 0.8380 & 0.4507\\
\bottomrule
\end{tabular}}
\end{table}

First, F1 is essentially flat across the four configurations at about 0.838, so the informative metric on POPE is precision rather than F1: the streams push the model towards being more conservative and more accurate, which shows up as a monotone rise in precision accompanied by a matching fall in recall and YesRatio. Second, the gain in precision is paid for in recall: the recall of C (0.7967) is below that of D (0.8260), and YesRatio drops from 0.4853 to 0.4507. This is a direct consequence of the design philosophy of DSCC, say nothing when nothing is clearly seen; the trade-off is acknowledged openly in Section~\ref{sec:4.4} and Section~\ref{sec:6} rather than sidestepped. The YesRatio of C nevertheless stays above 0.45, far from the red line of about 0.30 below which a model games the benchmark by refusing to answer, so the gain in precision comes from genuine grounding rather than from a degenerate answer distribution.

\subsubsection{CHAIR Hallucination in Open-Ended Captions}
\label{sec:4.2.2}

DSCC is first compared with decoding-time methods on CHAIR (Table~\ref{tab:6}), and the controlled four-way results follow (Table~\ref{tab:7}). All CHAIR runs use the same scorer eval\_chair\_official.py, max\_new\_tokens = 512 and the same prompt ``Please describe this image in detail.''; numbers from other papers are not quoted, since their CHAIR values differ in max\_new\_tokens, sampling scheme and image subset.

\begin{table*}[!t]
\centering
\caption{CHAIR-500 compared with decoding-time methods (same base model, same 500 images, same scorer, all reproduced under the unified protocol of this paper).}
\label{tab:6}
\footnotesize
\setlength{\tabcolsep}{5pt}
\begin{threeparttable}
\begin{tabular}{@{}llcccccc@{}}
\toprule
Method & Type & \#Words & Obj/Cap & CHAIR\_S$\downarrow$ & CHAIR\_I$\downarrow$ & Hall.\ obj./cap.\tnote{a}$\downarrow$ & Recall\\
\midrule
LLaVA-1.5 greedy & no intervention & 89.5 & 6.08 & 59.40 & 18.28 & 1.11 & 78.73\\
LLaVA-1.5 sampling & no intervention & 104.9 & 7.12 & 57.80 & 18.77 & 1.34 & 74.84\\
VCD & decoding-time & 104.0 & 7.61 & 58.60 & 17.06 & 1.30 & 79.31\\
OPERA & decoding-time & 93.1 & 7.41 & 45.20 & 13.27 & 0.98 & 79.57\\
DSCC (C, this paper) & training & 171.5 & 5.10 & \textbf{38.80} & \textbf{11.81} & \textbf{0.60} & 64.53\\
\bottomrule
\end{tabular}
\begin{tablenotes}[flushleft]\footnotesize
\item[a] Hallucinated objects per caption, Obj/Cap $\times$ CHAIR\_I, is the mean number of hallucinated objects in a caption; being absolute it must be read together with the Obj/Cap column.
\end{tablenotes}
\end{threeparttable}
\end{table*}

Reading this table requires distinguishing three kinds of metric. The first kind depends directly on object density: CHAIR\_S falls mechanically as the number of objects mentioned per caption falls, so the lowest CHAIR\_S in the table (38.80, DSCC) cannot by itself be taken as evidence of better grounding, since the Obj/Cap of DSCC (5.10) is indeed the lowest. The second kind is an absolute count: DSCC produces the fewest hallucinated objects per caption (0.60, against 0.98 for OPERA and 1.11 for greedy decoding). This quantity equals Obj/Cap $\times$ CHAIR\_I, however, and therefore remains proportional to object density; it cannot on its own answer the objection that hallucination was lowered by mentioning fewer objects. A conservative counterfactual: raising the object density of DSCC to OPERA's 7.41 while holding its CHAIR\_I fixed would give 0.88, still below OPERA's 0.98, but the margin would narrow from about 39\% to about 11\%. Only the third kind is genuinely independent of object density: precision per mention, 1 $-$ CHAIR\_I, normalises by the total number of mentions, and on this criterion DSCC is highest (88.19\% against OPERA's 86.73\%). Precision per mention, together with the D $\rightarrow$ C comparison of Section~\ref{sec:4.4.1}, is therefore taken as the primary evidence, while CHAIR\_S and hallucinated objects per caption are treated as auxiliary information to be read alongside Obj/Cap. DSCC and OPERA occupy two different operating points on the precision-recall frontier. DSCC trades 15 percentage points of object recall (64.53 against 79.57) for 1.46 percentage points of precision per mention; OPERA mentions more correct objects (6.43 against 4.50 per caption) at the cost of more hallucinated ones (0.98 against 0.60). Which is preferable depends on the application. No claim is made that DSCC dominates OPERA across the frontier; the claim is narrower, namely that within the long-caption regime, DSCC attains the highest precision per mention (Section~\ref{sec:4.3}), and that its conservative stance is a design choice rather than an evaluation loophole (Section~\ref{sec:6}).

\begin{table}[!t]
\centering
\caption{Four-way CHAIR-500 ablation (same 500 images, same scorer).}
\label{tab:7}
\footnotesize
\setlength{\tabcolsep}{3pt}
\resizebox{\ifdim\width>\columnwidth \columnwidth\else\width\fi}{!}{%
\begin{tabular}{@{}lccccc@{}}
\toprule
ckpt & CHAIR\_S$\downarrow$ & CHAIR\_I$\downarrow$ & Recall & \#Words & Obj/Cap\\
\midrule
D, both off & 41.60 & 12.42 & 65.76 & 170.1 & 5.22\\
A, perception only & 40.80 & 12.75 & 65.11 & 167.3 & 5.15\\
B, cognition only & 39.00 & \textbf{11.34} & 65.56 & 169.5 & 5.22\\
\textbf{C, full dual stream} & \textbf{38.80} & 11.81 & 64.53 & 171.5 & 5.10\\
\bottomrule
\end{tabular}}
\end{table}

The main model C attains the lowest sentence-level hallucination rate, CHAIR\_S = 38.80, which is 2.8 percentage points below the vanilla SFT baseline D (41.60). Crucially, this reduction happens without any compression of caption length: \#Words stays at roughly 170 across all four configurations and Obj/Cap at roughly 5.1-5.2, which shows that the dual-stream modules do not lower hallucination by the cheap route of shortening the output or mentioning fewer objects. They act on a fixed, verbose output regime already established by SFT. That property is exactly what the length-aware analysis of the next section builds on.

On the instance-level metric CHAIR\_I the best configuration is B (11.34) rather than C (11.81), a difference of 0.47. Every number reported here is a point estimate from a single evaluation run, with no interval estimate or significance test, so a difference of this magnitude is treated with consistent restraint: the 0.47 by which B beats C is neither dismissed as evaluation noise nor taken as evidence that C is inferior to B at the instance level. The gap is consistent with the more conservative object-extraction stance of C, in that adding the perception stream does not push instance-level precision any further, and it is reported faithfully in Section~\ref{sec:6} rather than dressed up as an advantage. A second source of caution is lexical: the official CHAIR vocabulary maps surface words to COCO categories, so the verb ``bearing'', the idiom ``a bird's eye view'' and the colour ``orange'' all count as objects. Such false positives are markedly more frequent in the more literary fine-tuned configurations than in the off-the-shelf baseline, a further reason to read differences of this size conservatively.

\subsection{Length-Aware Analysis}
\label{sec:4.3}

\begin{figure}[!t]
\centering
\includegraphics[width=1.0\columnwidth]{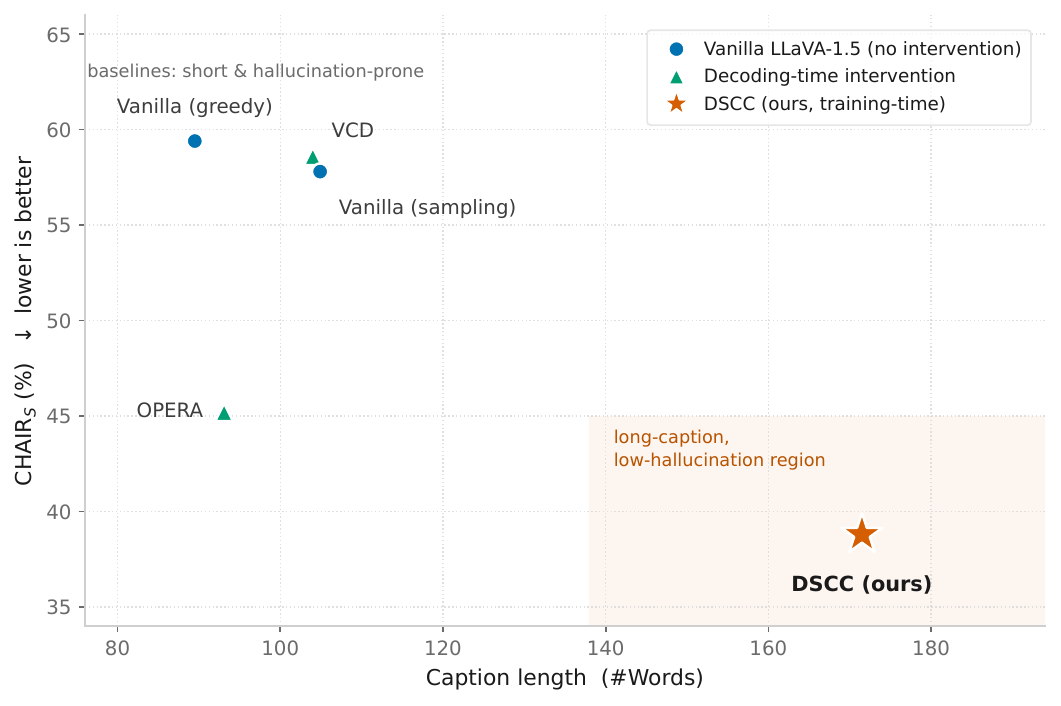}
\caption{The length-quality trade-off. Mean caption length (\#Words) against CHAIR\_S for every method, on the same 500 COCO images under the same scorer. The intervention-free and decoding-time baselines cluster in the short-caption region; DSCC is the only method that lands in the long-caption, low-hallucination region at the lower right.}
\label{fig:3}
\end{figure}

This section carries the new perspective of the paper. Existing methods have had their low hallucination verified mainly within the short-caption regime; whether their gains survive once a caption grows longer, so that object mentions and opportunities for error both rise, has so far lacked horizontal evidence gathered under one protocol with length and object density reported. Plotting the same-protocol data of Table~\ref{tab:6} on a single \#Words $\times$ CHAIR\_S plane (Fig.~\ref{fig:3}) makes the situation plain.

The three intervention-free and VCD points cluster in a small group at the upper left (roughly 90--105 words, CHAIR\_S 57--59\%). OPERA is the strongest baseline among them and takes the short-but-accurate route: it leaves caption length unchanged (93.1 words, comparable to the 89.5 of greedy vanilla) yet cuts CHAIR\_S to 45.20, and stops there. DSCC sits alone at the lower right (171.5 words, CHAIR\_S 38.80), the only method that pushes hallucination to its lowest value while lengthening the caption substantially, to about 1.9 times greedy vanilla, and it therefore does not lie on the established frontier along which longer captions bring more hallucination.

The figure also blocks the two most immediate objections to this work.

(1) ``CHAIR is low only because the captions are short.'' The \#Words of DSCC (171.5) is roughly 1.6 to 1.9 times that of every baseline (greedy 89.5, sampling 104.9, VCD 104.0, OPERA 93.1), so the objection fails on the data.

(2) ``CHAIR is lowered by mentioning fewer objects.'' This objection has to be met head on, because the Obj/Cap of DSCC (5.10) really is the lowest in the comparison. The answer is not denial but separation: precisely because object density cannot be controlled across methods, a control configuration D matched in both length and object density was trained (\#Words $\approx$ 170, Obj/Cap $\approx$ 5.2; Table~\ref{tab:7}). Under that same-length, same-density condition, C still lowers CHAIR\_S by 2.8 percentage points relative to D, and that portion of the gain cannot be explained by saying less; once length and object density are controlled, the most reasonable attribution is the dual-stream architecture. Removing the two confounders is not the same as removing sampling variance, however: the 2.8 points are a point estimate from a single evaluation without a significance test, so the attribution is directional and its certainty is not overstated.

\subsection{Ablation Study}
\label{sec:4.4}

This section decomposes the roles of the two streams and their synergy through a four-way ablation, and on that basis completes the two-level attribution on which the paper rests.

\subsubsection{Two-Level Attribution: Data Effect versus Net Architectural Gain}
\label{sec:4.4.1}

The standing of DSCC relative to the literature baselines has to be split into two levels, each measured against a different reference.

(1) First level, the contribution of the training paradigm (against the literature LLaVA-1.5): captions about 1.9 times longer with CHAIR\_S about 20 percentage points lower (59.40 $\rightarrow$ 38.80). It must also be said that this level comes with a wholesale shift in the shape of the output: object density falls from 6.08 to 5.22 and object recall from 78.73 to 65.76, so a substantial share of those 20 points should be attributed to mentioning fewer and more confident objects rather than to any improvement in grounding as such. The direct evidence is that D, vanilla SFT with both streams disabled, already produces 170 words with Obj/Cap 5.22 and CHAIR\_S 41.60. This paper therefore explicitly does not claim that the dual-stream modules make captions both longer and more accurate: the length comes from the SFT paradigm, and those 20 points do not belong to the architecture.

(2) Second level, the net gain of the dual-stream architecture (against vanilla SFT, D): on top of D, C adds CHAIR\_S $-$2.8, CHAIR\_I $-$0.61 and POPE Adversarial precision +3.3. That is the net architectural gain, real but modest. Its magnitude is reported honestly and is not inflated into a mechanistic breakthrough. Much of the literature on training-time hallucination mitigation reports only the total improvement over an off-the-shelf baseline without a vanilla-SFT control at the same corpus and step count, and therefore cannot tell how much of the reported gain actually came from the training data.

\subsubsection{Division of Labour and Synergy Between the Streams}
\label{sec:4.4.2}

Taking POPE Adversarial precision, CHAIR\_S and CHAIR\_I as three axes (values from Tables~\ref{tab:5} and~\ref{tab:7}), the configurations decompose as follows.

The cognition stream supplies the corrective, conservative capability by querying visual evidence at every step, which raises precision, and it is the main driver across metrics. The perception stream on its own makes the model more aggressive: configuration A has the highest YesRatio (0.5023) and the lowest precision, and that negative result matters in its own right, since pushing CLIP-style \cite{r30} contrastive learning down into the LLM does not automatically buy precision. Once it is stacked on the cognition stream, however, the two interact synergistically: the aggressive tendency of the fine-grained object anchors is held in check, and adversarial precision reaches the highest value among the four configurations (C = 0.8839).

This is deliberately described as an interaction, a synergy, rather than the simple addition of independent contributions: if the two streams merely added up, B $\rightarrow$ C would not exhibit the sign reversal whereby A alone is negative yet the combination is positive. On the main metrics (POPE Adversarial precision, Adversarial accuracy, CHAIR\_S) the ordering C \textgreater{} B \textgreater{} D holds and C \textgreater{} A holds, so the dual-stream design is justified by the ablation. Where a single configuration beats C it concedes elsewhere: B has the lowest CHAIR\_I (11.34) but a higher CHAIR\_S and a lower Adversarial precision (0.8638 against 0.8839), and D the highest POPE mean F1 (86.51) at 3.3 points less precision. C alone regresses on neither axis, which is what the dual stream buys. No claim is made, however, that both streams are indispensable or that the dual stream is universally optimal; the OOD results in the next section give that claim its precise boundary.

\subsection{Out-of-Domain Generalisation and Domain-Conditionality}
\label{sec:4.5}

The generalisation of the dual stream is examined on three OOD benchmarks.

\subsubsection{MME-Hallucination (Out of Distribution, Same Semantic Domain)}
\label{sec:4.5.1}

The ranking is C \textgreater{} B \textgreater{} A \textgreater{} D (Table~S-4). The main model C is best, improving on the vanilla SFT baseline D by 128.33 points, which is a clean out-of-domain win.

\subsubsection{HallusionBench (Charts: Genuinely Out of Domain)}
\label{sec:4.5.2}

The ranking is B \textgreater{} C \textgreater{} D \textgreater{} A (Table~S-5), and two conclusions follow. First, both C and B are clearly above D, so the generalisation brought by the dual stream, and especially by the cognition stream, still holds here. Second, and more tellingly, the main model C is not the best configuration on this benchmark: it is overtaken by B, cognition only.

\subsubsection{MMHal-Bench (Flickr Domain: A Null Result)}
\label{sec:4.5.3}

The four configurations are statistically indistinguishable (Table~S-6): C and D differ by about three questions, or 0.01 in mean score, which is less than one standard error. This is not presented as a win but reported faithfully as boundary evidence for domain-conditionality, with two caveats noted: the sample of N = 96 is too small, and the images come from Flickr and abstract scenes rather than from COCO.

\subsubsection{A Unified Insight: Predictable Domain-Conditionality}
\label{sec:4.5.4}

Collecting the in-domain result together with the three out-of-domain ones (Table~\ref{tab:12}) yields a claim more useful than universal superiority.

\begin{table}[!t]
\centering
\caption{Sign of the B $\rightarrow$ C synergy and the best configuration across benchmarks.}
\label{tab:12}
\footnotesize
\setlength{\tabcolsep}{3pt}
\resizebox{\ifdim\width>\columnwidth \columnwidth\else\width\fi}{!}{%
\begin{tabular}{@{}llcc@{}}
\toprule
benchmark & Image domain & synergy (B$\rightarrow$C) & Best configuration\\
\midrule
POPE (in-domain) & COCO & + & C\\
CHAIR (in-domain) & COCO & + (CHAIR\_S) & C\\
MME (OOD) & COCO object semantics & \textbf{+} & \textbf{C}\\
HallusionBench (OOD) & Charts and optical illusions & \textbf{$-$} & B\\
MMHal (OOD) & Flickr and abstract scenes & \textasciitilde{} (noise) & C (tied)\\
\bottomrule
\end{tabular}}
\end{table}

(1) The cognition stream is the domain-independent workhorse. It brings a positive gain on every benchmark, contributing the greater part of the +128 that D $\rightarrow$ C gains on MME, and it generalises stably.

(2) The perception stream is domain-conditional. The closer a benchmark is to COCO object semantics, the more useful it is: helpful on MME, which shares the semantic domain, and neutral or even harmful on HallusionBench, whose charts lie outside it. The reason is that the perception stream uses COCO class names explicitly as CLIP text anchors.

(3) The synergy holds only inside the competence domain of the perception stream. Within the COCO object domain (POPE, CHAIR, MME) C is best everywhere; once the data are genuinely out of domain, as with the charts of HallusionBench, the synergy breaks and B moves ahead.

The effectiveness of DSCC is accordingly stated in restricted form, namely effective within the COCO object semantic domain together with a predictable domain-conditionality, rather than universally optimal. Such a claim is falsifiable and checkable: given a new benchmark, deciding whether it falls inside the COCO object semantic domain is enough to predict whether the dual-stream synergy will hold.

\subsubsection{Qualitative Case Study}
\label{sec:4.5.5}

Qualitative comparisons on individual images are given in Figs.~S-1 and S-2 of the supplementary material, two cases per baseline. The texts quoted below are verbatim excerpts from the CHAIR-500 generations, with omissions marked [\ldots]; baseline A is the off-the-shelf model and baseline B the length- and density-matched configuration D.

(1) Against baseline A, \#508218 closes with ``a dog located towards the right side of the image'' although the annotation lists only zebra, car and person, and \#569674, annotated with zebra alone, fills an empty sky with ``several birds scattered throughout the scene''; DSCC describes both scenes with no hallucinated object at all.

(2) Against baseline B, \#300855 seats the crowd on furniture that is not there, ``others are sitting on benches nearby'', where DSCC describes the same gathering without them; and \#258433, whose annotation contains a truck and no car, is miscategorised rather than invented, ``a food truck \ldots\ while a blue car is parked nearby'', where DSCC gives the annotated category, ``a blue truck is parked by the sidewalk''.

The second case is a category correction rather than an invented object, corroborating the mechanistic account of Section~\ref{sec:4.4.2}, under which the perception anchors are responsible for fine-grained category discrimination.

\subsection{Methodological Implications of the Length-Aware Perspective}
\label{sec:4.6}

The length-aware analysis (Section~\ref{sec:4.3}) also exposes a structural confounder in hallucination evaluation: because CHAIR-style metrics are coupled to generation length and object density, a method can lower CHAIR\_S simply by mentioning fewer objects. This is not an accusation: under the unified protocol the object densities and recalls of VCD and OPERA are no lower than the intervention-free baseline's, so neither gamed the metric by saying less. The confounder constrains this work first of all: the Obj/Cap of DSCC, 5.10, is the lowest in the comparison.

Two methodological recommendations follow. First, comparisons among hallucination mitigation methods should report, and where possible align, generation length and object density, at minimum as mandatory \#Words and Obj/Cap columns; otherwise a lower CHAIR may be no more than a shorter output in disguise. Second, any such method fine-tuned on detail-rich data should report a data-only control in the same configuration; otherwise its architectural contribution cannot be audited.

\section{Conclusion and Future Work}
\label{sec:6}

This paper has studied object-level hallucination in MLLMs. With control configuration D matched in both length and object density, the increase in caption length is attributed chiefly to the SFT data paradigm rather than to the dual-stream architecture, whose net gain is real but limited. The value of this work lies not in state-of-the-art numbers, none of which are claimed, but in a new perspective, a complete empirical account and a predictable boundary of validity.

The boundaries deliberately exposed indicate where subsequent research can act. The full model is conservative, saying nothing when nothing is clearly seen: its POPE recall of 0.797 is below the 0.826 of control configuration D, and against OPERA it trades 15 percentage points of object recall for higher precision per mention. On the instance-level metric it is not strictly best either: B (11.34) beats C (11.81) by 0.47. The remaining limitations fall into three groups. In capability, the synergy is bound hard to COCO object semantics and breaks, or even becomes a burden, once a task is genuinely out of domain; logical hallucination, consistent with the visible objects yet violating spatial, counting or commonsense relations, receives no explicit signal. In method, the reading that the cognition stream severs the path along which deep reasoning drifts from the visual evidence rests on no rigorous causal analysis, and the perception and cognition layers follow design considerations rather than a layer-by-layer sweep, so the combination cannot be asserted optimal. In evaluation, every difference reported is a point estimate from a single run without significance testing, and the out-of-domain scoring protocol differs from the official leaderboards, so those absolute scores are not comparable, MMHal-Bench returning a null result that serves only as boundary evidence for domain-conditionality. The plans that follow are accordingly: a confidence threshold adjustable at inference; open-vocabulary text anchors in place of fixed class names; training and evaluation repeated over multiple random seeds with confidence intervals; and a DPO \cite{r23} stage stacked on the policy obtained by DSCC training, the two being complementary.

\end{document}